\documentclass[preprint,12pt]{elsarticle}

\usepackage{amssymb}
\usepackage{amsmath}
\usepackage{booktabs}
\usepackage{algorithm}
\usepackage{multirow}
\usepackage{CJK}

\journal{Arxiv}

\begin{document}

\begin{frontmatter}



\title{Agents on the Bench: Large Language Model Based Multi Agent Framework for Trustworthy Digital Justice}


\author{Cong Jiang} 
\author{Xiaolei Yang} 

\affiliation{organization={Peking University},
            addressline={No.5, Yiheyuan Road, Haidian District}, 
            city={Beijing},
            postcode={100871}, 
            country={China}}

\begin{abstract}
The justice system has increasingly employed AI techniques to enhance efficiency, yet limitations remain in improving the quality of decision-making, particularly regarding transparency and explainability needed to uphold public trust in legal AI. To address these challenges, we propose a large language model based multi-agent framework named AgentsBench, which aims to simultaneously improve both efficiency and quality in judicial decision-making. Our approach leverages multiple LLM-driven agents that simulate the collaborative deliberation and decision making process of a judicial bench. We conducted experiments on legal judgment prediction task, and the results show that our framework outperforms existing LLM based methods in terms of performance and decision quality. By incorporating these elements, our framework reflects real-world judicial processes more closely, enhancing accuracy, fairness, and society consideration. AgentsBench provides a more nuanced and realistic methods of trustworthy AI decision-making, with strong potential for application across various case types and legal scenarios. 

\end{abstract}




\end{frontmatter}



\section{Introduction}
\label{sec1}

The judiciary system has seen significant developments in recent years, with the increasing use of artificial intelligence techniques. AI can help streamline case management, improve consistency in rulings, and make legal systems more accessible. \cite{bellandi2024entity} Despite these benefits, current AI research for judicial applications faces several challenges. Many AI models developed for tasks such as legal judgment prediction focus heavily on accuracy metrics, but they often suffer from biases, lack explainability, and fail to consider ethical and social effect of algorithm judgment. \cite{cui2023survey} Moreover, these techniques usually rely on a single model to make decision, which are far from real-world judicial decision-making processes, such as discussion, debate and consensus-building among judges. \cite{xu2024through} Our work aims to address these issues by proposing a solution that not only improves decision quality but also enhances the transparency of digital justice system, improving their reliability and applicability.

The use of a deliberative bench composed of multiple judges is crucial in judicial systems,. This bench, often consisting of a mix of professional judges and sometimes lay jurors, plays an essential role in maintaining fairness and minimizing individual biases. The collaborative nature of these deliberations ensures that multiple viewpoints are considered before reaching a final decision, thereby increasing the reliability of judicial outcomes. \cite{dawson1960history}  Early work on agent-based models (ABM) simulation attempted to introduce a multi-agent system in the legal field. \cite{benthall2021agent} However, the ABM approach is difficult to simulate legal process consisting of complex natural language discussion and legal concepts.

Large language model (LLM) based multi-agent systems have recently demonstrated remarkable potential in various domains by effectively tackling complex challenges through collaborative problem-solving and cognitive diversity. \cite{abdelnabi2023llm} Unlike conventional ABM, which face challenges in capturing complex legal discussions due to difficulties with natural language processing and understanding intricate legal concepts, LLMs possess the unique ability to generate and interpret language, making them more suitable for simulating deliberative processes. Building on these successes, we propose applying an LLM-based multi-agent framework to the legal domain to enhance fairness, depth, and quality in judicial decision-making. 

In this paper, we introduce a LLM-based multi-agent framework named \textbf{AgentsBench} to simulate the collaborative nature of a judicial bench. Our framework consists of multiple LLM-driven agents, each representing different judicial roles, who deliberate collectively to form a decision. By incorporating discussion, debate, and consensus-building among these agents, our approach better reflects the actual judicial process, thereby increasing both accuracy, tranparancy and explainbability. The AgentsBench framework combines the capabilities of LLMs with the structured deliberation of real judicial environments, ultimately enhancing the quality and trustworthiness of AI-supported digital justice.

\section{Related Work}\label{sec2}

\subsection{Large Language Modes in Law}\label{subsec1}

AI applications in the legal domain have seen significant advancements with the development of  LLMs. These models show capability in tackling various legal scenarios, such as legislation, legal literacy, and justice.  For instance,  \cite{liga2023fine} demonstrated that GPT-3 can effectively classify legal and deontic rules, such as obligations, permissions, and constitutive rules, even with limited data, outperforming prior models on the same task. \cite{jiang2024leveraging} presented a novel application of LLMs to improve legal literacy for non-experts through storytelling.  \cite{deroy2024applicability} explored the applicability of  LLMs for summarizing lengthy and complex legal case judgments. Jiang and Yang explored legal syllogism prompt engineering to improve the legal reasoning performance of LLMs. \cite{jiang2023legal}. Some work focused on how to train the language model into a domain-specific LLM for law,  to make it better fit the requirement of legal domain.\cite{licari2022italian} fine-tuned  BERT model on Italian legal data to improve NLP tasks within the Italian legal domain. \cite{huang2023lawyer} explored adapting LLMs to the legal domain via pre-training and supervised fine-tuning, while mitigating hallucination by integrating a retrieval module for relevant legal articles.   While these works highlight significant progress in using LLMs for legal tasks, there remain gaps with the real world justice system. Current LLM-based research often work on specific tasks but lack the exploration of complexity and dynamics of justice and legal system. \cite{hamilton2023blind} trained nine separate models with the respective authored opinions of each supreme court judges, shedding light on our research. But it only gives the results of the judges' votes, and doesn't model the complex decision making process.

\subsection{LLM-based Multi-agent System}\label{subsec2}

LLM-based multi-agent systems have recently shown great potential in a variety of domains. \cite{park2023generative} By allowing multiple agents to work collaboratively, these frameworks leverage cognitive diversity and interaction to solve complex tasks more effectively than single LLM systems. \cite{he2023lego} In general-purpose contexts, LLM-based multi-agent frameworks have improved problem-solving capabilities by utilizing the combined reasoning and knowledge of multiple agents. \cite{hongmetagpt} In specialized domains such as healthcare, systems like Agent Hospital have demonstrated the benefits of using multiple LLM-driven agents for simulating hospital environments and facilitating medical decision-making processes. \cite{li2024agent} These systems have consistently achieved superior outcomes compared to standalone LLMs by fostering dynamic interaction and shared learning. \cite{qian2024chatdev} Inspired by such advancements, our work aims to explore the application of an LLM-based multi-agent framework in the legal domain, using multiple agents to simulate collegial bench, enhancing the fairness, depth, and quality of justice decision making. 

\subsection{Legal Judgment Prediction}\label{subsec3}

The task of Legal Judgment Prediction is originally defined to predict the results of a legal judgment given the descriptions of a fact. \cite{kort1957predicting}. Earlier research focused on collecting legal case datasets for different jurisdictions and improve deep learning algorithms on them. For example,  \cite{xiao2018cail2018}  introducing large-scale Chinese legal data, CAIL2018, for predictive tasks involving charges, penalties, and relevant articles.   \cite{chalkidis2019neural} built an English LJP dataset that contains cases from the European Court of Human Rights. \cite{luo2017learning} proposes an attention-based neural network that jointly models charge prediction and relevant article extraction. \cite{hong2023improving} introduces PekoNet, a framework that integrates abstractive text summarization to improve the predictive accuracy of LJP models for colloquial case descriptions.  \cite{chien2024legal} develops TWLJP, a dataset of indictments for judgment prediction, and improves charge prediction through interactive message passing and prompt-based learning, benefiting prosecutors in detecting discrepancies and managing legal knowledge. Recently, more work has begun to reflect on the value of the above approaches. \cite{an2022charge} proposes a new framework to assess whether LJP models conform to legal theories, revealing gaps between law and techniques in existing models. \cite{medvedeva2023legal} argued that many LJP studies do not adequately address practical needs, emphasizing the importance of explainability, user-centric approaches, and proper data usage for reliable LJP systems. While performance is important, interpretability and transparency are critical for trustworthy legal AI systems. Most existing studies lack adequate reasoning behind their predictions, which limits their practical applicability.

\section{Agents on the bench}
This section outlines the components of our proposed framework. \textit{AgentsBench} uses LLM based model based agents to simulate the complex dynamics of a collegial bench in the judicial decision-making process. The framework consists of four primary stages: (i) Bench Selection, (ii) Independent Sentencing, (iii) Deliberation, and (iv) Final Decision Making. Figure 1 provides an overview of the AgentsBench process.

\begin{figure}[t]
\centering
\includegraphics[width=0.9\textwidth]{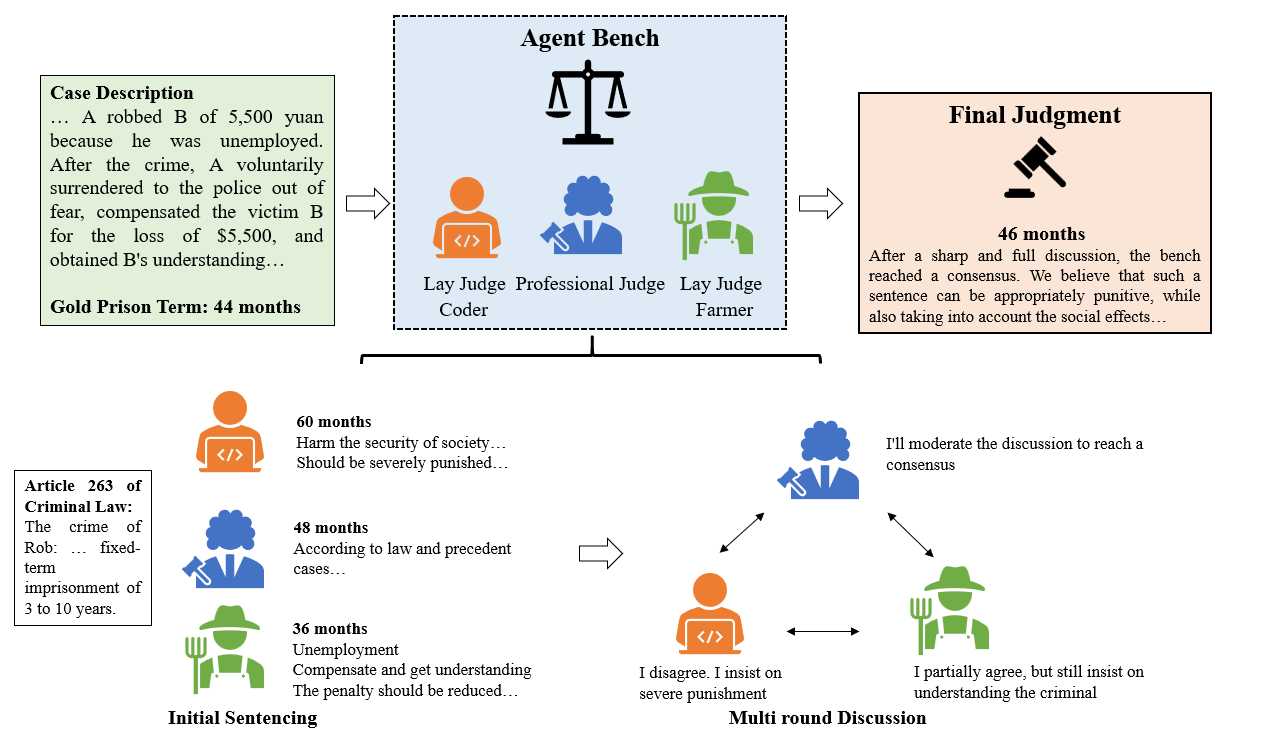}
\caption{Overview of the AgentsBench Framework.
The figure illustrates the framework simulating a judicial decision-making process. It features an agent bench consisting of two lay judges and one professional judge. Each agent initially proposes an independent sentencing decision based on the case details. Subsequently, the agents engage in multi-round deliberation, moderated by the professional judge, to reconcile differing perspectives and achieve consensus. This collaborative process reflects the essence of AgentsBench, leveraging diverse viewpoints to reach a balanced and fair final judgment, which considers both legal standards and social effects.} \label{fig1}
\end{figure}

\subsection{LLM Agent}

Each agent within the framework is driven by a LLM and is designed to mimic the different actors present in a typical judicial bench. Inspired by other work on agent simulation[cite] we set each agent to have certain generic capabilities that empower them to operate autonomously during simulations. Specifically, the agents are capable of: (1) \textbf{Planning}: Each agent formulates an initial plan based on the given context, deciding what actions to take and whether to recall the memorized experience; (2) \textbf{Acting}: The agents execute their plans, contributing to the deliberation by presenting arguments, questioning each other, and providing insights. (3) \textbf{Reflecting}: After each round of deliberation, agents reflect on their contributions, learning from their observations and from the input of other agents. (4) \textbf{Memory}: Agents can store encountered precedents, allowing them to summarize past experiences and apply them to similar future cases.

To closely simulate a real-world collegial bench, we set up different types of agents, each fulfilling a distinct judicial role. Specifically, we designed judge agents and juror agents to replicate the collaborative judicial process. The \textbf{judge agents} are responsible for overseeing the entire decision-making process, moderating discussions, summarizing arguments, and ultimately facilitating or making a consensus-based final decision. This role is guided by a specific prompt that encourages moderation, evaluation, and adherence to legal standards. In contrast, \textbf{juror agents} act as lay judges, providing perspectives that reflect community values and societal norms. Juror agents actively participate in discussions, contributing to the collective decision-making process.  The prompts for judge and juror agents are generated using the LLM to ensure coherence and to provide different focuses for each role. For example, prompts for jurors emphasize societal and ethical considerations, while prompts for judges emphasize adherence to legal principles and moderation.

\subsection{Bench Selection}

In this initial stage, AgentsBench forms a collegial bench by selecting agents to represent both professional judges and lay judges. The composition of the bench is carefully designed to reflect the diversity present in real-world judicial decision-making processes. To achieve this, a professional judge with extensive legal expertise is assigned as the presiding judge, providing a central figure with deep knowledge of legal standards and procedures. In addition, a diverse pool of lay judge agents is maintained, each representing different backgrounds, areas of expertise, and societal perspectives. Lay judges are then randomly selected from this pool for each simulation, ensuring that the resulting bench is dynamic and reflective of the varied experiences found in real-world judicial panels. This selection process helps simulate the interplay between professional legal reasoning and the broader viewpoints contributed by lay members of the bench.

\subsection{Independent Sentencing}

Once the collegial bench is formed, each agent independently evaluates the case and proposes an initial sentence. During this stage, all agents—including both professional and lay judges—thoroughly review the case details, applying their unique backgrounds, perspectives, and knowledge to form an individual understanding of the situation.

Mathematically, we denote the initial sentencing decisions as:

\[
S^{(0)} = \{s_1^{(0)}, s_2^{(0)}, \dots, s_n^{(0)}\}
\]

where \( s_i^{(0)} \) represents the initial sentencing decision made by agent \( i \). Each agent \( i \) derives \( s_i^{(0)} \) based on their independent analysis of the case, which can be described by the function:

\[
s_i^{(0)} = A_i(C, P_i)
\]

Here:

\begin{itemize}
    \item \( C \) represents the case details provided to all agents.
    \item \( P_i \) denotes the unique personal factors of agent \( i \), including their background, experience, and perspectives.
    \item \( A_i \) is the analysis function used by agent \( i \), which takes into account both \( C \) and \( P_i \) to generate the initial sentencing decision \( s_i^{(0)} \).
\end{itemize}

Importantly, each agent documents the rationale behind their decision, capturing their interpretation of the law as well as their subjective views on the specifics of the case. This process ensures that the diversity of perspectives among the agents is preserved, forming the set \( S^{(0)} \), which serves as the foundation for the subsequent deliberation stages.

\subsection{Deliberation}

During each round of deliberation, all agents independently update their sentencing decisions based on the deliberation outcomes. The update function for agent \(i\) in round \(t+1\) can be represented as:

\[
s_i^{(t+1)} = U(s_i^{(t)}, D^{(t)})
\]

where \(s_i^{(t)}\) is the sentencing decision of agent \(i\) in the previous round, and \(D^{(t)}\) contains the discussion content from round \(t\). The function \(U\) prompts each agent to reconsider their initial sentencing, incorporating new arguments and perspectives shared by others. This process allows each agent to either modify their decision or provide further justification.

After each round, the presiding judge evaluates whether the bench has reached a consensus. Instead of relying on a strict numerical threshold for agreement, the presiding judge—an LLM agent—uses its interpretative capabilities to determine whether the agents’ updated positions converge sufficiently in terms of content and rationale. This is represented by:

\[
C^{(t)} = \text{JudgeEval}(S^{(t+1)}, D^{(t)})
\]

where \(C^{(t)}\) is a binary indicator of consensus (\(C^{(t)} = 1\) if consensus is achieved, otherwise \(C^{(t)} = 0\)). The presiding judge’s evaluation considers not only the similarity of the updated decisions \(S^{(t+1)}\) but also the coherence and convergence of arguments presented during discussion \(D^{(t)}\).

The deliberation proceeds iteratively through rounds until consensus is reached or the maximum allowed number of rounds is completed. This adaptive approach enables the presiding judge to assess agreement dynamically, reflecting the fluid and nuanced nature of human judicial decision-making.

\subsection{Decision Making}  

In the concluding stage, the presiding judge synthesizes the discussion outcomes to reach a final decision. This process can be described as follows:

Once all rounds of deliberation are complete, the presiding judge agent analyzes the various points raised during the discussions. This analysis includes evaluating arguments, identifying recurring themes, and integrating the perspectives of both professional and lay judges. If a consensus was reached during the deliberation, the presiding judge ratifies that consensus as the final sentencing decision. However, if significant disagreements persist, the presiding judge must weigh all contributions and utilize their expertise to determine an appropriate final judgment.

Mathematically, the final sentencing decision \( S_{\text{final}} \) can be represented as:

\[
S_{\text{final}} = g(S^{(T)}, D_{\text{history}})
\]

where:

\begin{itemize}
    \item \( S^{(T)} \) represents the set of updated sentencing decisions from all agents after the final round of deliberation.
    \item \( D_{\text{history}} \) includes the complete history of discussion points raised during all deliberation rounds.
    \item \( g \) is the synthesis function, executed by the presiding judge to combine all available input and formulate the final sentencing decision.
\end{itemize}

Finally, the presiding judge provides a comprehensive justification for the final judgment, incorporating insights gathered from each round of deliberation. This justification captures both the rational analysis of legal principles and the subjective viewpoints contributed by different bench members, ensuring transparency and thoroughness in the decision-making process.

The AgentsBench framework aims to capture the nuanced interplay between professional legal knowledge and diverse societal perspectives in judicial decision-making. By simulating this complex process, AgentsBench provides a novel approach to studying the impact of bench composition on sentencing outcomes and the dynamics of judicial deliberations.

\section{Experiment}

\subsection{Task and dataset}

We chose Prison Term Prediction task to evaluate the ability of our framework for judicial decision making. Prison Term Prediction is the task of predicting the potential criminal sentence of the defendant based on given case facts and legal provisions. This task can be viewed as a subtask of legal judgment prediction.[cite] In criminal legal judgment, the court determines the article of law, charge and prison term based on the facts of the case. Determining the article of law and charge is largely based on the knowledge and application of law, leaving judges and bench with limited room for discretion. In contrast, Prison Term Prediction requires deciding the appropriate length of a prison term, which offers the bench significantly more discretionary space. In civil law countries, the criminal law typically provides a range rather than a fixed term for each charge. For example, in cases of less severe intentional homicide in China, the prison term is from 3 to 10 years, specified by Article 232 of the Chinese Criminal Law. This broader range of discretion in sentencing allows judges to consider factors such as the severity of the crime, mitigating circumstances, and the defendant’s background, making it an ideal task to evaluate the nuanced decision-making abilities of our methodology. Hence, we select the Prison Term Prediction task as it provides a more comprehensive assessment of the framework's capability to support the judicial decision-making process, especially in scenarios where balancing various factors and applying judicial discretion are crucial. 

The Prison Term Prediction task can be formally defined as follow. The input are the fact description of a case \textbf{X} with the applicable legal article \textbf{l} and charge \textbf{c}. Since prison term prediction is the final step in criminal judgment, we assume that the articles and charges have already been determined. Therefore they are used as inputs along with the facts of the case. The task is to predict the potential length of the prison term, represented as \(y = n\). In the AgentsBench framework, the decision-making process involves a collegial bench, composed of multiple agents including both professional and lay judges. Each agent independently analyzes the input \((\mathbf{X}, l, c)\) and proposes an initial sentence \(s_i^{(0)}\). The collective deliberation process of the bench then refines these initial proposals to arrive at a final sentencing decision \(y\).

\[
y = f(\mathbf{X}, l, c)
\]

For our experiments, we selected evaluation data from LawBench dataset, which is a benchmark dataset designed to assess the legal capabilities of LLMs in Chinese.[cite] Specifically, LawBench includes various legal AI tasks, among which Tasks 3-5 is the Prison Term Prediction task. The data of tasks 3-5 is based on the Chinese AI and Law challenge dataset (CAIL2018) \cite{xiao2018cail2018}. It is a Chinese criminal case dataset widely used for LJP research. They are all real cases collected from China Judgments Online, the official  website  that  publishes  cases  and  decisions  from  Chinese  courts. Each case involves both factual description and legal judgment. The judgment further includes three parts: legal articles, charges, and prison terms. The LawBench remove out cases involving the death penalty and a life sentence in prison, and then randomly choose 500 cases as the test dataset. To assist the LLMs perform the prison time prediction task, it added the charge name and the full content of applicable article at the end of the fact description when creating the test dataset.

The evaluation results of numerous LLMs has been publicly shared by the LawBench, such as GPT4, GPT 3.5,  and Qwen, offering an objective baseline reference for our research. This enables us to benchmark our AgentsBench framework against existing models in a fair and transparent manner. 

\subsection{Setup}

\subsubsection{Baselines}
We compare our framework with the following baselines to evaluate its performance :

• Standard Prompt: This baseline involves prompting  LLMs to output only the prison term, without any additional contextual information or step-by-step reasoning. This approach aims to measure the basic decision making capability of LLMs without specialized prompting strategies  

• CoT: The zero-shot Chain of Thought (CoT) method enhances reasoning by incorporating the phrase "Let's think step by step" into the prompt. This method encourages the LLM to engage in a reasoning process before providing an answer, potentially improving the quality and accuracy of the prediction by prompting the model to articulate intermediate reasoning steps \cite{kojima2022large}. 

• LS: Legal syllogism (LS) prompting is a zero-shot approach that instructs the LLM to apply syllogistic reasoning to legal judgment prediction tasks. The prompt first defines the legal syllogism structure and then guides the model through applying it to the given case.\cite{jiang2023legal} This approach tests the model's ability to logically derive the outcome based on legal articles. By comparing against LS prompting, we can evaluate how effective our multi-agent deliberation process is compared to a purely logical and structured method. 

Each of these baselines represents a different aspect of LLM capabilities, ranging from basic prediction to advanced logical reasoning and formal legal reasoning. These comparisons provide insight into the performance of our multi-agent framework relative to other established approaches in the field of legal judgment prediction.

\subsubsection{Implementation} 
We used the close-source model GPT-4, GPT-3.5-Turbo and open-source Chinese model Qwen-7B for all experiments. All experiments were conducted in a zero-shot setting, where no specific training examples were provided to the models beforehand. To ensure reproducibility, we set the temperature to 0 and the top\_p to 1.0 for all prediction steps.   We made right truncation to the input of some cases, where the fact description exceeds the length limitation.

\subsection{Evaluation} 

The evaluation is divided into two parts: Performance Evaluation and Quality Evaluation. The first part focuses on the quantitative metrics that assess the accuracy of predictions made by these methods, while the second part involves qualitative assessments made by automated or human evaluators to assess the legality, rationality, and morality considerations of the outcomes. 

\subsubsection{Performance Evaluation }

The performance evaluation assesses the accuracy of the predicted outcomes generated by different methods compared to the gold standards. Since the model outputs often include extra details like reasoning and explanations, we extract the numeric prison term from the LLM output and convert it into a standard format. First, Chinese numbers are converted to Arabic numerals. Then, we extract the values before time units month and year, converting every term into months. The ground truth labels are also standardized to months to ensure consistency in evaluation. 

We evaluate the prison term prediction task using the normalized log distance (nLog-distance) as the scoring metric. It is used to capture the continuity of sentence lengths. First, we calculate the logarithm of the absolute difference between the predicted term and the gold standard prison term. We then normalize this value to fall within the range of 0 to 1, using the following formula: 

\begin{equation}
    score_  = \frac{\log(|\text{predicted term} - \text{gold answer}| + 1)}{\log(\text{maximum possible difference} + 1)}
\end{equation}

\subsubsection{Quality Evaluation }

Given the intricate nature of legal decision making, it is challenging to rely solely on automated metrics for a full assessment of the framework's capabilities. Therefore, we also conducted a human evaluation to qualitatively assess the deliberations and final decisions of these methods. For this evaluation, we employed a panel of three legal professionals who independently reviewed a random sample of 100 case outputs. 

The evaluators assessed each case based on three criteria: \textbf{legality}, \textbf{logicality}, and \textbf{morality}. Legality is evaluated by determining whether the decisions complied with relevant legal articles and legal theory. Logicality is judged based on the coherence and logical progression of arguments provided by the method. Morality is assessed to ensure that some moral, ethical and social factors are properly taken into account in the decision-making process. Each criterion was rated as either "True" (1) or "False" (0) to capture whether the method's output met the expected standard, and inter-rater reliability was measured using Cohen's kappa to evaluate agreement among evaluators.

\subsection{Results}

\begin{table}[h]
\centering
\caption{Comparison of Different Methods on Prison Term Prediction Task}
\label{tab:detailed_results}
\resizebox{\textwidth}{!}{
\begin{tabular}{llcccc}
\hline
\textbf{Model} & \textbf{Method} & \textbf{Performance (\%)} & \textbf{Legality (\%)} & \textbf{Logicality (\%)} & \textbf{Morality (\%)} \\
\hline
\multirow{4}{*}{Qwen}     
     & Standard Prompt       & 74.22 &       &       &       \\
     & CoT                   & 73.32 & 49.3  & \textbf{53.6} & 49.2 \\
     & LS                    & 72.82 & 53.2  & 52.5  & 48.9 \\
     & AgentsBench           & \textbf{78.25} & \textbf{55.4} & 53.4  & \textbf{68.7} \\
\hline
\multirow{4}{*}{GPT-3.5} 
     & Standard Prompt       & 75.13 &       &       &       \\
     & CoT                   & 76.45 & 51.0  & 55.0  & 50.1 \\
     & LS                    & 74.72 & 54.8  & 54.4  & 49.6 \\
     & AgentsBench           & \textbf{80.81} & \textbf{55.3} & \textbf{55.2} & \textbf{72.1} \\
\hline
\multirow{4}{*}{GPT-4} 
     & Standard Prompt       & 80.98 &       &       &       \\
     & CoT                   & 79.76 & 54.1  & 57.3  & 51.5 \\
     & LS                    & 80.28 & \textbf{56.8} & 56.6  & 52.2 \\
     & AgentsBench           & \textbf{86.33} & 56.5  & \textbf{58.2} & \textbf{76.2} \\
\hline
\end{tabular}
}
\end{table}

Table 1 compares different methods on the Prison Term Prediction task across three models: Qwen, GPT-3.5, and GPT-4. The evaluation is based on \textbf{performance}, \textbf{legality}, \textbf{logicality}, and \textbf{morality}.

AgentsBench consistently outperformed other methods, achieving the highest scores across all three models. Specifically, GPT-4 with AgentsBench reached \textbf{86.33\%}, significantly higher than other methods, highlighting the strength of the LLM agent-based deliberation framework. Interestingly, CoT and LS methods did not always yield better results compared to direct output (Standard Prompt); in some cases, these methods even led to decreased performance, which confirms findings from other studies. \cite{blair2023can} We believe that this is related to the specificity of prison-term prediction. Unlike charge prediction, which relies more on legal rules and logic, prison term is not exclusively determined by law and logic. In contrast, the AgentsBench approach did not harm performance. This emphasizes the effectiveness of structured multi-agent deliberation in enhancing predictive accuracy without the risks associated with other prompting techniques. 

Where AgentsBench truly excelled was in \textbf{morality}. It achieved a score of \textbf{76.2\%} with GPT-4, significantly outperforming other methods. This highlights the strength of the multi-agent deliberative approach in achieving ethically balanced outcomes, which are crucial in legal decision-making processes that require careful consideration of fairness, goodness, and social justice. This advantage reflects AgentsBench's ability to integrate diverse perspectives to produce morally robust and balanced decisions.

In terms of \textbf{legality} and \textbf{logicality}, AgentsBench performed competitively compared to baseline methods. While other frameworks, like Legal Syllogism (LS), showed slightly higher scores in these dimensions, AgentsBench still demonstrated solid adherence to legal principles, achieving \textbf{56.5\%} in legality and \textbf{58.2\%} in logicality with GPT-4. This indicates that AgentsBench remains consistent in delivering reliable legal and logical evaluations.
 

\subsection{Case Study}
To further evaluate the performance of our AgentsBench framework, we conducted a detailed case study focused on a complex bribery and fraud case.Details of the case and a full bench diliberation content are in the appendix. 

In this case, we can see the sentencing deliberation process where multiple agents, including professional judges and lay jurors, evaluated the facts of the case. Initially, each agent independently proposed a sentencing decision. The presiding judge A proposed a harsher sentence of \textbf{60 months} based on the severity of Liu’s offenses, emphasizing the need for a strong deterrent. In contrast, Judge B proposed \textbf{48 months}, citing mitigating factors such as the defendant’s remorse and the fact that this was a first offense. Lay juror C suggested \textbf{54 months}, balancing the severity of the crime with the defendant's remorse.

The deliberative rounds demonstrated the collaborative strength of the AgentsBench framework. During the initial round of discussions, the presiding judge summarized the various perspectives, and the agents collectively discussed key factors, including the social impact of the crime and the defendant’s repentance. After thorough debate, a consensus was reached in the second round, with all agents agreeing on a \textbf{54-month} sentence. Considering the gold label is \textbf{58months}, the bench results has been very close to gold. This outcome illustrated how a deliberative approach facilitates nuanced decision-making that takes into account diverse viewpoints while balancing legal rigor and societal considerations. The collaborative discussions led by multiple agents effectively captured the complexities of the case. The presiding judge acted as a moderator, synthesizing the input from the other agents, which enabled a more rounded assessment compared to individual agent evaluations.

The final decision balanced the need for punishment with a recognition of the defendant's remorse and the desire to provide a rehabilitative opportunity. Legal experts rated the deliberation process highly in terms of its \textbf{reasoning quality}, \textbf{logical consistency}, and \textbf{legality}, as the agents applied the relevant provisions from Articles 266 and 385 of the Chinese Criminal Law appropriately and coherently. The agents’ use of statutory law, combined with careful consideration of the defendant’s background and mitigating factors, led to a final judgment that was both legally sound and ethically considerate.

\section{Conclusion}

In this paper, we presented AgentsBench, an LLM-based multi-agent framework that simulates judicial decision-making by incorporating deliberative discussions among multiple agents. Our findings show that AgentsBench not only improves the accuracy of legal judgment predictions but also enhances fairness and ethical considerations in legal decision-making processes. Our framework is highly extensible and can be adapted to other types of legal cases and broader judicial scenarios, providing a pathway for more comprehensive and realistic applications of AI in the justice system.

\bibliographystyle{elsarticle-num} 
\bibliography{main}



\appendix
\section{Case Detail}
\textbf{Case Detail:} 
From January 2012 to July 2013, while serving as the director of the Second Office of the Land and Resources Bureau of Yuexi County, the defendant Liu A was assigned by the Yuexi County Land and Resources Bureau to work as a staff member at the Ruicheng Home Plaza Command, Zhongzhou Road Command, and Ma'ao Abandoned Soil Field, where he was engaged in land acquisition investigation, expropriation, and assistance in demolishing house measurements and compensation. He took advantage of his position to accept 28,900 yuan in property from others and seek benefits for others. His behavior constituted ××. In 2012, the defendant Liu A was entrusted by Wang A, the general manager of Yuexi County Ruicheng Real Estate Development Co., Ltd., to negotiate with Wang, who operated the "Dawanzhuang Restaurant", on the demolition of the "Dawanzhuang Restaurant". The defendant Liu A fabricated facts, concealed the truth, and defrauded Wang A of 300,000 yuan, which he kept for himself. His behavior also constituted ××, and he should be punished for several crimes. I hereby file a public prosecution and request a sentence in accordance with the law. I also present and read out relevant witness testimony and defendant's confession and other evidence in court.

Crime: Fraud; Bribery. 

Legal Articles: Articles 266, 385. 
Article 266: If a person defrauds public or private property in a relatively large amount, he shall be sentenced to fixed-term imprisonment of not more than three years, criminal detention or public surveillance and shall be fined or fined alone; if the amount is huge or there are other serious circumstances, he shall be sentenced to fixed-term imprisonment of not less than three years but not more than ten years and shall be fined; if the amount is particularly huge or there are other particularly serious circumstances, he shall be sentenced to fixed-term imprisonment of not less than ten years or life imprisonment and shall be fined or have his property confiscated. If otherwise provided in this Law, the provisions shall prevail.
Article 385: If a state functionary uses his position to solicit property from others, or illegally accepts property from others to seek benefits for others, he shall be guilty of bribery. If a state functionary, in economic dealings, violates state regulations and accepts kickbacks and fees of various names for personal use, he shall be punished for bribery.

\section{Bench Discussion Content} 

\subsection*{Presiding Judge: Zhou}
\begin{quote}
\textbf{Sentence Term:} 60 months

\textbf{Reason:} Based on the specific circumstances of the case, defendant Liu, during his tenure as a public official, took advantage of his position to accept bribes of 28,900 yuan, and, while negotiating demolition work, fabricated facts to defraud 300,000 yuan. He is guilty of bribery and fraud. According to Articles 266 and 385 of the Criminal Law, the fraud involved a large sum and serious circumstances, and the bribery involved abusing official power, which is a serious offense.

During sentencing, I considered the circumstances of the crime, the amount involved, and its impact on society. After comprehensive analysis, I believe a heavier sentence is warranted. While the defendant may have shown some degree of remorse, the nature of his crime and the harm caused to society make a heavy penalty necessary to serve as a warning and deterrent. Therefore, sentencing him to 60 months in prison is appropriate.
\end{quote}

\subsection*{Judge: Zhang }
\begin{quote}
\textbf{Sentence Term:} 48 months

\textbf{Reason:} During his tenure as a public official, defendant Liu used his position to accept bribes totaling 28,900 yuan, and while negotiating demolition work, fabricated facts to defraud 300,000 yuan. This constitutes bribery and fraud. According to Articles 266 and 385 of the Criminal Law, the amounts involved were substantial, and the circumstances were serious, making this a significant legal offense.

When determining the sentence, I took into account the following factors: Firstly, the social harm caused by the defendant's actions is considerable, especially given that he was a state official whose behavior undermined public interest and societal trust. Secondly, while the defendant may have shown some remorse, the severity and magnitude of his crimes necessitate an appropriate level of punishment. Lastly, given that the defendant might be a first-time offender who has shown remorse, a moderately lenient punishment could be considered.

Therefore, after considering all these factors, I believe a sentence of 48 months in prison is appropriate, serving both punitive and deterrent purposes while recognizing the defendant's expression of remorse.
\end{quote}

\subsection*{Lay Judge: Su}
\begin{quote}
\textbf{Sentence Term:} 54 months

\textbf{Reason:} Defendant Liu, during his tenure as a public official, abused his position to accept bribes of 28,900 yuan and fabricated facts to defraud 300,000 yuan, which constitutes bribery and fraud. These actions not only harmed public interests but also undermined public trust in the government, thus having significant social harm.

In sentencing, I believe the following factors should be considered: Firstly, the nature of the defendant's actions was serious. As a state official, using his position for illegal activities warrants severe punishment to uphold the authority of the law and social justice. Secondly, while the defendant may have expressed some remorse, the magnitude and severity of his crimes still warrant appropriate punishment as a warning.

Therefore, sentencing him to 54 months in prison is appropriate as it not only serves as a punishment for his actions but also takes into account his remorse, striving for a balance between punishment and rehabilitation.
\end{quote}

\subsection*{Presiding Judge Deliberation (Round 1): Zhou}
\begin{quote}
Presiding Judge (Zhou):

Colleagues, thank you for your input. Now, let's discuss the differences in sentencing opinions. First, I'll summarize each perspective:

1. \textbf{Presiding Judge Zhou}: The defendant's actions were serious, involving substantial bribery and fraud, with significant social harm. Therefore, I recommend a sentence of 60 months in prison to serve as a warning and deterrent.

2. \textbf{Judge Zhang}: Although the defendant's actions were serious, he may have shown some remorse and is a first-time offender. Therefore, I recommend a sentence of 48 months in prison to acknowledge his expression of remorse.

3. \textbf{Lay Judge Su}: Supports a heavier sentence, noting that the defendant's actions harmed public interests and suggesting 54 months in prison to uphold the authority of the law and social justice.

In our discussion, we need to focus on the following aspects:

- \textbf{Nature of the crime and social impact}: Does the defendant's abuse of power for bribery and fraud warrant a heavier sentence?
- \textbf{Remorse and first-time offense}: Has the defendant shown remorse, and should this be considered in sentencing?
- \textbf{Justice and social effects of sentencing}: How should we balance the punishment of the defendant with the social deterrent effect?

I invite each of you to express your views on these points, particularly regarding whether the defendant's remorse is sufficient to affect sentencing. Let us explore this perspective further to see if we can reach a consensus.
\end{quote}

\section*{Sentencing and Deliberation Process - Continued}

\subsection*{Judge Deliberation (Round 1): Zhang}
\begin{quote}
In this case, there are differences among the members of the collegial panel regarding the sentencing of defendant Liu. As a judge, I will analyze the legal application, the nature of the crime, the social impact, and the defendant's remorse to better explain my viewpoint.

Firstly, regarding the defendant's actions, according to Articles 385 and 266 of the Criminal Law, his conduct constitutes bribery and fraud. Liu, during his tenure as a public official, accepted bribes of 28,900 yuan and fraudulently obtained 300,000 yuan by fabricating facts. The amounts involved are substantial, and the circumstances are severe. Such conduct not only harms public interests but also damages public trust in the government, leading to significant social harm.

Secondly, when determining the sentence, we must consider the circumstances of the crime, the amount involved, the social impact, and the defendant's remorse. It is important to note that while the defendant's actions are serious, he has shown genuine remorse and has no prior criminal record. These factors warrant a more lenient approach to sentencing, reflecting both his willingness to take responsibility and the need to balance punishment with opportunities for rehabilitation.

The Presiding Judge Zhou's suggestion of 60 months emphasizes deterrence and the need to protect societal interests, and I acknowledge the validity of these concerns. However, I believe that we must also give weight to the defendant's remorse and the mitigating factors related to his first-time offense.

Lay Judge Su's suggestion of 54 months strikes a certain balance between punishment and rehabilitation, but I believe that a lighter sentence could still effectively convey the seriousness of the offense while recognizing the defendant's mitigating factors.

In conclusion, considering all viewpoints, I maintain that a sentence of 48 months is appropriate. This decision takes into account the need for punishment, deterrence, and the defendant's genuine remorse, providing a more balanced approach to justice. A 48-month sentence would still serve as an adequate punishment while also allowing for the possibility of rehabilitation, ultimately aligning with the relevant provisions of the Criminal Law and ensuring a just outcome. 

Therefore, I recommend that the panel reach a consensus on imposing a sentence of 48 months on the defendant Liu.
\end{quote}

\subsection*{Lay Judge Deliberation (Round 1): Su}
\begin{quote}
As a Lay Judge, I believe that the sentencing in this case is highly important as it concerns both social justice and public trust in the law. Liu's actions are not only violations of the law but also serious infringements on public interests. As a public official, he abused his position for bribery and fraud, which undermined public trust in the government. The social harm of such behavior is evident.

The Presiding Judge Zhou proposed a sentence of 60 months, which I understand emphasizes deterrence and punishment. Such a sentence can convey a zero-tolerance stance on similar behavior, uphold the authority of the law, and strengthen public confidence. As a public servant, Liu 's actions were not only personal misconduct but also a betrayal of public office, and society's tolerance for such behavior should be very low.

Judge Zhang's suggested sentence of 48 months, although considering the defendant's remorse, does not fully reflect the seriousness of the behavior. The amounts and circumstances involved indicate the severity of his actions, and a lenient sentence could lead to a misunderstanding of the law, implying that illegal behavior can easily evade punishment.

Therefore, I maintain that sentencing Liu to 54 months in prison is appropriate. This is not only a punishment for his actions but also a warning to society, effectively upholding the authority of the law and public trust. As judges and jurors, we must reflect our firm stance on social justice in our sentencing.
\end{quote}

\subsection*{Presiding Judge Consensus Determination: Zhou}
\begin{quote}
\textbf{Conclusion: No}

\textbf{Main Points of Disagreement:} The panel members have significant differences regarding sentencing. Presiding Judge Zhou suggests a sentence of 60 months, emphasizing the severity of the defendant's actions and the need for deterrence. Judge Zhang suggests 48 months, considering the defendant's remorse and first-time offense. Lay Judge Zhang supports 54 months, aiming to strike a balance between punishment and rehabilitation. The different perspectives on sentencing and the rationale behind them have led to a lack of consensus.
\end{quote}

\subsection*{Presiding Judge Updated Sentencing: Zhou }
\begin{quote}
Presiding Judge (Zhou):

After the panel discussion, I believe we need to reassess the sentencing for defendant Liu . While I initially proposed a sentence of 60 months, emphasizing the severity of his actions and the deterrent effect on society, after hearing the views of the judges and jurors, I realize that we must more comprehensively consider the defendant's remorse and first-time offense.

Firstly, the defendant's actions indeed constitute bribery and fraud, with substantial amounts involved and significant social harm. However, Judge Zhang Jun pointed out the importance of considering the defendant's remorse and first-time offense. If the defendant sincerely showed remorse during the trial and it was his first offense, this could be considered grounds for a lighter sentence.

Secondly, Lay Judge Zhang Wei's suggestion of a 54-month sentence strikes a balance between punishment and rehabilitation. I believe that further adjustments can be made to better reflect the defendant's remorse and first-time offense.

Therefore, I have decided to modify my sentencing opinion to 54 months. This sentence can reflect the punishment for the defendant's actions while considering his remorse and first-time offense to some extent. Such a decision aligns with the relevant provisions of the Criminal Law and conveys a firm stance on social justice while maintaining the authority of the law.

In conclusion, I propose that the panel supports sentencing defendant Liu to 54 months. I invite the judges to discuss this opinion.
\end{quote}

\subsection*{Judge Revises Sentence - Zhang}
\begin{quote}
After the in-depth discussion among the collegial panel, I have re-examined the sentencing of defendant Liu and considered the opinions and suggestions presented by all parties. Based on the following reasons, I have decided to revise my sentencing recommendation to 54 months of imprisonment.

Firstly, while the defendant Liu's actions constitute bribery and fraud, involving a substantial amount and causing significant social harm, we must also consider his remorse and the fact that this is his first offense when determining the sentence. Judge Zhang emphasized the defendant's remorse, particularly his attitude during the trial, which should be an important factor considered during sentencing. If the defendant can demonstrate sincere remorse and actively return the illicit gains, these can be considered as grounds for a lighter punishment.

Secondly, Lay Judge Su proposed a sentence of 54 months, which strikes a balance between punishment and education, effectively serving as a deterrent for similar behavior while recognizing the defendant's remorse. This sentencing is in line with the relevant provisions of the Criminal Law, while also conveying a firm stance on social justice and upholding the authority of the law.

Finally, considering that Liu was a public official, the harm caused to the public interest cannot be ignored. Therefore, the severity of the punishment must still be reflected in the sentencing. A sentence of 54 months not only effectively punishes the defendant but also acknowledges his remorse to a certain extent, balancing punishment and rehabilitation.

In conclusion, I believe that sentencing defendant Liu to 54 months of imprisonment is appropriate. This decision not only maintains the authority of the law but also helps promote the defendant's reform, maximizing the societal effect. I hope the collegial panel can reach a consensus on this opinion.
\end{quote}

\subsection*{Lay Judge Revises Sentence - Su}
\begin{quote}
As a lay judge, after the collegial panel's discussion, I have re-evaluated the sentencing of defendant Liu. Based on the following reasons, I have decided to revise my sentencing recommendation to 54 months of imprisonment.

Firstly, the defendant Liu's actions indeed constitute bribery and fraud, involving a significant amount and causing serious social harm. Such behavior not only harms public interests but also undermines public trust in the government. Therefore, it is necessary to impose a severe penalty on the defendant to uphold the authority of the law and social justice.

Secondly, in determining the sentence, we must consider the defendant's remorse and the fact that this is his first offense. Although his actions caused harm to society, his genuine remorse during the trial and being a first-time offender can be grounds for leniency. Judge Zhang and Lay Judge Su both mentioned the defendant's remorse, which I believe is significant. We must strike a balance between punishment and education, ensuring that we punish the defendant's actions while also giving him a chance to reform.

Lastly, a sentence of 54 months strikes a balance between punishment and education, effectively deterring similar offenses while acknowledging the defendant's remorse. This sentencing is not only in line with the relevant provisions of the Criminal Law but also conveys a firm stance on social justice while upholding the authority of the law.

In conclusion, I believe sentencing defendant Liu to 54 months is appropriate. This decision not only effectively punishes his actions but also acknowledges his remorse to some extent, achieving a balance between punishment and rehabilitation. I hope the collegial panel can reach a consensus on this opinion.
\end{quote}

\subsection*{Collegial Panel Discussion (Round 2) - Zhou}
Presiding Judge (Zhou):
\begin{quote}

Colleagues, thank you for your thorough discussion. We have reached a certain consensus regarding the sentencing of defendant Liu, and the current recommendation is 54 months of imprisonment. I will now summarize the opinions of each party and guide us further to ensure we reach a final agreement on the sentencing.

1. **Presiding Judge Zhou**: Initially suggested 60 months, emphasizing the severity of the defendant's actions and their deterrent effect on society. However, after considering the opinions of all parties, I recognized the importance of remorse and the fact that this was a first offense, and revised my recommendation to 54 months.

2. **Judge Zhang**: Believes the defendant's remorse and first-time offense should be considered, supporting the 54-month sentence as a balance between punishment and education.

3. **Lay Judge Su**: Also supports the 54-month sentence, emphasizing the social harm caused by the defendant's actions while recognizing the importance of remorse. Su believes that a 54-month sentence effectively serves as a deterrent.

Based on this, I would like to further discuss the following issues to help us reach a final consensus:

- **Specifics of the Remorse**: Did the defendant demonstrate specific remorse during the trial? Did he actively return the illicit gains? Are these factors sufficient to influence sentencing?

- **Evaluation of Social Impact**: Have we adequately considered the long-term impact of the defendant's actions on society? Is a 54-month sentence sufficient to convey a zero-tolerance attitude towards similar actions?

- **Consistency in Legal Application**: Are there similar sentencing standards in comparable cases? Can we ensure that this decision maintains consistency in the application of the law?

Please share your thoughts on the above issues, especially regarding whether the defendant's remorse is sufficient to affect the sentencing. Let's explore this further to see if we can reach a consensus.
\end{quote}

\subsection*{Collegial Panel Discussion (Round 2) - Zhang}
\begin{quote}
In this case, the members of the collegial panel have reached a certain consensus regarding the sentencing of the defendant Liu. The current agreed-upon opinion is to sentence Liu to 54 months of imprisonment. As a judge, I will analyze the legal application, the nature of the crime, the social impact, and the defendant's remorse to better explain my viewpoint.

Firstly, Liu's conduct constitutes bribery and fraud, involving a substantial amount, causing significant social harm. According to Articles 385 and 266 of the Criminal Law, both bribery and fraud involve large sums and severe circumstances, deserving corresponding punishment. The defendant used his position to engage in illegal activities, harming public interests and undermining trust in the government, which clearly shows the social harm of his actions.

Secondly, when determining the sentence, we must comprehensively consider the circumstances of the crime, the amount involved, the social impact, and the defendant's remorse. While the defendant may have shown some remorse, the nature of his crime and its impact on society necessitate a significant penalty. Presiding Judge Zhou initially proposed a 60-month sentence, emphasizing the deterrent and punitive effects on society, which is crucial. As a public official, Liu's conduct was not just an individual violation of the law, but a serious betrayal of public duties. The tolerance for such behavior in society should be extremely low.

However, Judge Zhang and Lay Judge Su have suggested a sentence of 54 months, taking into account the defendant's remorse and first-time offense, which are indeed factors that should be considered during sentencing. Whether the defendant has genuinely shown remorse during the trial and whether he actively returned the illicit gains are elements that can be grounds for leniency.

After the collegial panel's discussion, I believe that 54 months of imprisonment strikes an appropriate balance between punishment and rehabilitation, effectively conveying a warning to others while acknowledging the defendant's remorse. Such a sentence not only aligns with the relevant provisions of the Criminal Law but also conveys a firm stance on upholding social justice while maintaining the authority of the law.

In conclusion, I support sentencing Liu to 54 months of imprisonment. This sentence effectively punishes the defendant's actions while acknowledging his remorse to some extent, achieving a balance between punishment and rehabilitation. I hope the collegial panel can reach a consensus on this opinion.
\end{quote}

\subsection*{Collegial Panel Discussion (Round 2) - Su}
\begin{quote}
As a lay judge, I believe that the sentencing of defendant Liu in this case is not only about his personal fate but also about social justice and the authority of the law. After a thorough analysis of the case, I maintain that sentencing Liu to 54 months of imprisonment is appropriate, for the following reasons:

Firstly, Liu's conduct constitutes bribery and fraud, involving substantial amounts and significant social harm. As a public official, he abused his position to engage in illegal activities, damaging public interests and undermining public trust in the government. The social harm caused by such behavior is evident, and severe punishment is necessary to uphold the authority of the law and social justice.

Secondly, during sentencing, we must consider the defendant's remorse and first-time offense. Although his actions have harmed society, his genuine remorse during the trial and the fact that this was his first offense can be considered as grounds for leniency. Judge Zhang and Presiding Judge Zhou both mentioned the defendant's remorse, which I believe should be taken seriously. We need to strike a balance between punishment and education, ensuring that we effectively punish the defendant while giving him the opportunity to reform.

Finally, a sentence of 54 months strikes a balance between punishment and rehabilitation, effectively serving as a deterrent for similar offenses while recognizing the defendant's remorse. This sentence is not only consistent with the relevant provisions of the Criminal Law but also conveys a firm stance on upholding social justice while maintaining the authority of the law.

In conclusion, I believe that sentencing Liu to 54 months of imprisonment is appropriate. This decision not only effectively punishes his actions but also acknowledges his remorse to some extent, achieving a balance between punishment and rehabilitation. I hope the collegial panel can reach a consensus on this opinion.
\end{quote}

\subsection*{Presiding Judge's Consensus Decision - Zhou}

\begin{quote}
Conclusion: Yes

The final sentencing opinion is: defendant Liu is sentenced to 54 months of imprisonment.

After thorough discussion among the collegial panel, all three members have reached a consensus, agreeing that 54 months of imprisonment is appropriate for effectively punishing Liu's conduct while also acknowledging his remorse to some extent, achieving a balance between punishment and rehabilitation.

\end{quote}

\subsection*{Summary of Collegial Panel Discussion - Zhou}
\begin{quote}
In the trial of this case, the members of the collegial panel conducted an in-depth discussion on the sentencing of defendant Liu and ultimately reached a consensus, sentencing him to 54 months of imprisonment. The following is a summary of the discussion process and the reasons for reaching a consensus:

\#\#\# Summary of the Discussion Process:

1. **Initial Opinions**:
   - Presiding Judge Zhou initially recommended 60 months, emphasizing the severity of the defendant's actions and their deterrent effect on society.
   - Judge Zhang proposed 48 months, considering the defendant's remorse and first-time offense.
   - Lay Judge Su supported a sentence of 54 months, seeking a balance between punishment and education.

2. **Analysis of Differences**:
   - There were differences in the basis and perspective of sentencing. The presiding judge emphasized social impact and deterrence, while the judge focused on remorse and first-time offense considerations.

3. **Revisiting Sentencing**:
   - After the collegial panel's discussion, Presiding Judge Zhou recognized the importance of the defendant's remorse and first-time offense, and revised his recommendation to 54 months.
   - Judge Zhang and Lay Judge Su also supported the 54-month sentence, believing it struck a balance between punishment and rehabilitation.

4. **Final Consensus**:
   - During the second round of discussion, all members of the collegial panel expressed support for a 54-month sentence, believing that such a decision effectively punishes the defendant while recognizing his remorse.

\#\#\# Reasons for Reaching a Consensus:

1. **Severity of the Crime**:
   - Liu's conduct constitutes bribery and fraud, involving a significant amount, harming public interests, and undermining public trust in the government.

2. **Consideration of Remorse**:
   - The defendant demonstrated a degree of remorse during the trial, and since this is his first offense, leniency should be considered during sentencing.

3. **Consistency in Legal Application**:
   - A sentence of 54 months is consistent with the provisions of the Criminal Law, upholding the authority of the law while conveying a firm stance on social justice.

4. **Balance Between Punishment and Education**:
   - Sentencing Liu to 54 months effectively serves as a deterrent for similar offenses while recognizing the defendant's remorse, achieving a balance between punishment and rehabilitation.

In conclusion, after thorough discussion, the members of the collegial panel believe that sentencing Liu to 54 months of imprisonment is appropriate, effectively punishing his actions while also acknowledging his remorse to some extent.
\end{quote}

\label{app1}










\end{document}